\documentclass[conference,a4paper]{IEEEtran}

\usepackage{cite}
\usepackage{graphicx}
\usepackage{subfig}
\usepackage{multirow}
\usepackage{graphicx}
\usepackage[table,xcdraw]{xcolor}
\usepackage[absolute,showboxes]{textpos}

\TPMargin{5pt}

\newcommand{\copyrightstatement}{
    \begin{textblock}{0.85}(0.08,0.90)    % tweak here: {box width}(leftposition, rightposition)
         \noindent
         \footnotesize
Copyright 2019 IEEE. Published in the Digital Image Computing: Techniques and Applications, 2019 (DICTA 2019), 2-4 December 2019 in Perth, Australia. Personal use of this material is permitted. However, permission to reprint/republish this material for advertising or promotional purposes or for creating new collective works for resale or redistribution to servers or lists, or to reuse any copyrighted component of this work in other works, must be obtained from the IEEE. Contact: Manager, Copyrights and Permissions / IEEE Service Center / 445 Hoes Lane / P.O. Box 1331 / Piscataway, NJ 08855-1331, USA. Telephone: + Intl. 908-562-3966.
    \end{textblock}
}

\begin{document}
%
% paper title
% can use linebreaks \\ within to get better formatting as desired
\title{Automatic Weight Estimation of\\ 
Harvested Fish from Images}

% for over three affiliations, or if they all won't fit within the width
% of the page, use this alternative format:
% 
% \author{\IEEEauthorblockN{11-July-2019: Submitted to DICTA-2019 for double-blinded peer review}}

\author{\IEEEauthorblockN{Dmitry A. Konovalov\IEEEauthorrefmark{1},
Alzayat Saleh\IEEEauthorrefmark{1},
Dina B. Efremova\IEEEauthorrefmark{2},
Jose A. Domingos\IEEEauthorrefmark{1}, 
Dean R. Jerry\IEEEauthorrefmark{1}
}
\IEEEauthorblockA{\IEEEauthorrefmark{1}College of Science and Engineering,
James Cook University,
Townsville, 4811, Australia\\ 
dmitry.konovalov@jcu.edu.au, alzayat.saleh@my.jcu.edu.au, 
jose.domingos1@jcu.edu.au, dean.jerry@jcu.edu.au}
\IEEEauthorblockA{\IEEEauthorrefmark{2}
Funbox Inc.,
Moscow, Russian Federation\\ 
dina.efremova85@gmail.com}
% {This version was submitted for review in July-2019 and accepted for presentation at DICTA-2019, 2-4 December 2019.\\ 
% Final version maybe slightly different.}
}
% 

% author names and affiliations
% use a multiple column layout for up to three different
% affiliations

%%%%%  Ready for submission
% \makeatletter
% \newcommand{\linebreakand}{%
%   \end{@IEEEauthorhalign}
%   \hfill\mbox{}\par
%   \mbox{}\hfill\begin{@IEEEauthorhalign}
% }
% \makeatother
% \author{\IEEEauthorblockN{
% Dmitry A. Konovalov}
% \IEEEauthorblockA{College of Science and Engineering\\
% James Cook University,
% Townsville, Australia\\
% dmitry.konovalov@jcu.edu.au}
% \and
% \IEEEauthorblockN{Alzayat Saleh}
% \IEEEauthorblockA{College of Science and Engineering\\
% James Cook University,
% Townsville, Australia\\
% alzayat.saleh@my.jcu.edu.au}
% % \and
% \linebreakand % <------------- \and with a line-break
% \IEEEauthorblockN{Jose A. Domingos}
% \IEEEauthorblockA{College of Science and Engineering\\
% James Cook University,
% Townsville, Australia\\
% jose.domingos1@jcu.edu.au}
% \and
% \IEEEauthorblockN{Dean R. Jerry}
% \IEEEauthorblockA{College of Science and Engineering\\
% James Cook University,
% Townsville, Australia\\
% dean.jerry@jcu.edu.au}
% }

% use for special paper notices
% \IEEEspecialpapernotice{}

% make the title area
\maketitle
% \copyrightnotice
\copyrightstatement

\begin{abstract}
\boldmath
Approximately 2,500 weights and corresponding images of harvested 
\textit{Lates calcarifer} (Asian seabass or barramundi) were collected at 
three different locations in Queensland, Australia. 
Two instances of the LinkNet-34 segmentation Convolutional 
Neural Network (CNN) were trained. 
The first one was trained on 200 manually segmented fish masks with excluded fins and tails. 
The second was trained on 100 whole-fish masks. 
The two CNNs were applied to the rest of the images and yielded automatically segmented masks. 
The one-factor and two-factor simple 
mathematical weight-from-area models 
were fitted on 1072 area-weight pairs from the first two
locations, where area values were extracted from the automatically segmented masks. 
When applied to 1,400 test images 
(from the third location), the one-factor whole-fish mask model 
achieved the best mean absolute percentage error (MAPE), $\mbox{MAPE}=4.36\%$.
Direct weight-from-image regression CNNs were also trained, 
where the no-fins based CNN performed best on the test images 
with $\mbox{MAPE}=4.28\%$.

\end{abstract}
% IEEEtran.cls defaults to using nonbold math in the Abstract.
% This preserves the distinction between vectors and scalars. However,
% if the conference you are submitting to favors bold math in the abstract,
% then you can use LaTeX's standard command \boldmath at the very start
% of the abstract to achieve this. Many IEEE journals/conferences frown on
% math in the abstract anyway.

% no keywords

% For peer review papers, you can put extra information on the cover
% page as needed:
% \ifCLASSOPTIONpeerreview
% \begin{center} \bfseries EDICS Category: 3-BBND \end{center}
% \fi
%
% For peerreview papers, this IEEEtran command inserts a page break and
% creates the second title. It will be ignored for other modes.
\IEEEpeerreviewmaketitle

% no \IEEEPARstart
\section{Introduction}
% no \IEEEPARstart

Economic competition, large volumes of animals, and increasing human labor cost drive the development and deployment of computer vision (CV) systems within the aquaculture 
industry \cite{hong2014visual,miranda2017prototype,sanchez2018automatic,zion2000vivo,saberioon2017application}. As an example, a CV system could automatically measure or estimate fish morphological features (length, width, and mass) 
\cite{balaban2010prediction,viazzi2015automatic,zenger2017next,domingos2014fate,konovalov2018estimating,konovalov2018automatic} 
on an industrial scale through an automated process. 
While the fish length (or any other visible sizes) can be estimated directly from the imagery 
\cite{miranda2017prototype,konovalov2018automatic,konovalov2017ruler}, 
the fish mass $M$ can only be approximately inferred
\cite{balaban2010prediction,viazzi2015automatic,konovalov2018estimating}.
Note that even the fish length extraction from images remains an active area of research \cite{Monkman2019length}.
Hereafter, the terms \textit{mass} and \textit{weight} were used interchangeably and as equivalent within the context of the out-of-water harvested fish.

The most commonly used approach to weight estimation uses fish length $L$ as 
a predictor 
variable \cite{sanchez2018automatic}. For example,
Sanchez-Torres \textit{et al}. \cite{sanchez2018automatic} 
estimated $L$ from fish ({\em Orechromis niloticus})  
contour $C$ and then treated 
fish mass $M$ as a response variable:
\begin{equation}
\label{eq:1a}
L=f(C), \ \ \ M=g(L),
\end{equation}
using five different mathematical and machine learning models, 
where $f$ and $g$ denote such models in general sense.
The best performing models were 3rd degree 
polynomials for both the $L=f(C)$ length-from-contour and $M=g(L)$ mass-from-length estimators.
When fitted on 75 images, the first half of the available images, and 
then tested on the second half of the images, the 3rd degree polynomial models  
achieved the mean absolute percentage errors (MAPE) $3.6\%$ in the length estimations
and $\mbox{MAPE}=11.2\%$ in the weight predictions.

Viazzia \textit{et al}. \cite{viazzi2015automatic} worked with a dataset of 120 measurements of
Jade perch ({\em Scortum barcoo}) covering $29-491$g mass range. Three mathematical models
were considered:
\begin{equation}
\label{eq:1b}
\mbox{Polynomial:} \ \ M=a + bS+cL+dH,
\end{equation}
\begin{equation}
\label{eq:1c}
\mbox{Linear:} \ \ M=a + bS,
\end{equation}
\begin{equation}
\label{eq:1d}
\mbox{Power curve:} \ \ M=aL^b,
\end{equation}
where $S$ was fish body surface area (with or without fins) and $H$ was fish height.
When tested on 64 images not used in the models' fitting process, the polynomial model 
(Eq.~\ref{eq:1b}) was the 
best performing model attained $\mbox{MAPE}=5\%$ from fish contours with or without fins.
Since only the length $L$ was used in \cite{sanchez2018automatic} as a feature variable, 
the only comparable model (in \cite{viazzi2015automatic}), which also used $L$ solely, 
was the power-curve model (Eq.~\ref{eq:1d}) that 
achieved $\mbox{MAPE}=10\%$ for contours without fins and $\mbox{MAPE}=12\%$ with fins.
Therefore, the 3rd degree polynomial model ($\mbox{MAPE}=11.2\%$) 
from \cite{sanchez2018automatic} was consistent and comparable
with the results ($\mbox{MAPE}=10-12\%$) of the power-curve model (Eq.~\ref{eq:1d})
from \cite{viazzi2015automatic}. 

Observing that Viazzia \textit{et al}. \cite{viazzi2015automatic} 
reported $\mbox{MAPE}=5-6\%$ using only 
the surface area $S$ (Eq.~\ref{eq:1c}), 
Konovalov \textit{et al}. \cite{konovalov2018estimating} fitted 
the following two mathematical models for harvested Asian seabass (\textit{Lates calcarifer}, also known as barramundi in Australia): 
\begin{equation}
\label{eq:1}
M=c S^{3 / 2}, \ \ c=0.170,
\end{equation}
\begin{equation}
\label{eq:2}
M=a S^{b}, \ \ a=0.124, \ \  b=1.55,
\end{equation}
where the mass $M$ was measured in grams and the fish body surface 
area $S$ was in cm\(^2\) for images with the scale of 1~mm-per-pixel. 
MAPE values were 5.1\% and 4.5\% for 
the single-factor (Eq.~\ref{eq:1}) and two-factor (Eq.~\ref{eq:2}) models, respectively, 
when fitted on 1072 different fish images from two different barramundi farms (Queensland, Australia) \cite{domingos2014fate}. 
In general, the fitting parameters $a$, $b$, and $c$ are 
species-dependent \cite{Huxley1924,ZION2012125}.

\begin{figure}[htbp]
\begin{center}
\includegraphics[width=0.48\textwidth]
{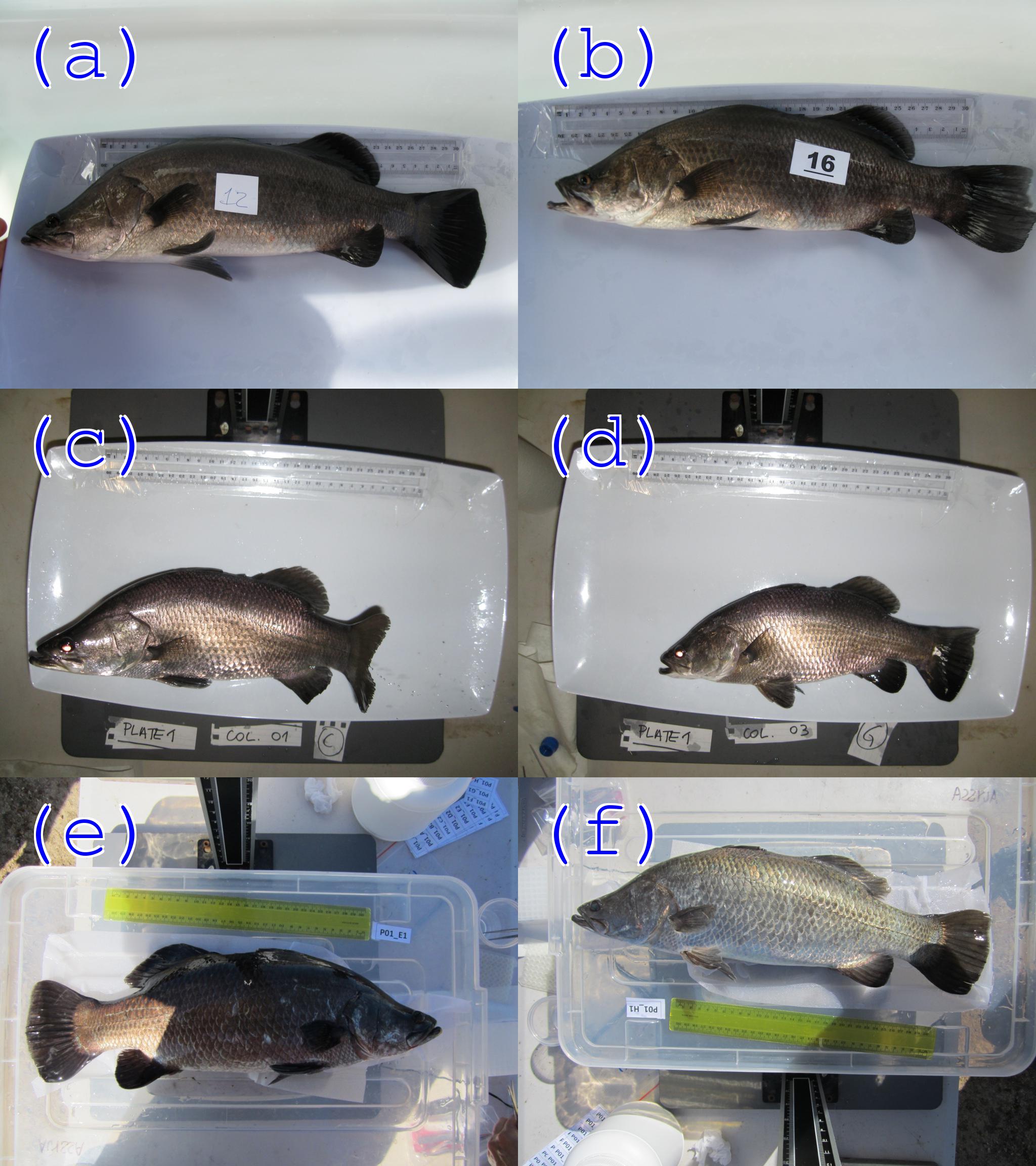}
% {./figures/fig1dk.eps}
\end{center}
\caption{Samples of original images from the 
used datasets: BR445 (a) and (b), BW1400 (c) and (d), BA600 (e) and (f).}
\label{fig:fig1}
\end{figure}

\begin{figure}[htbp]
\begin{center}
\includegraphics[width=0.48\textwidth]
{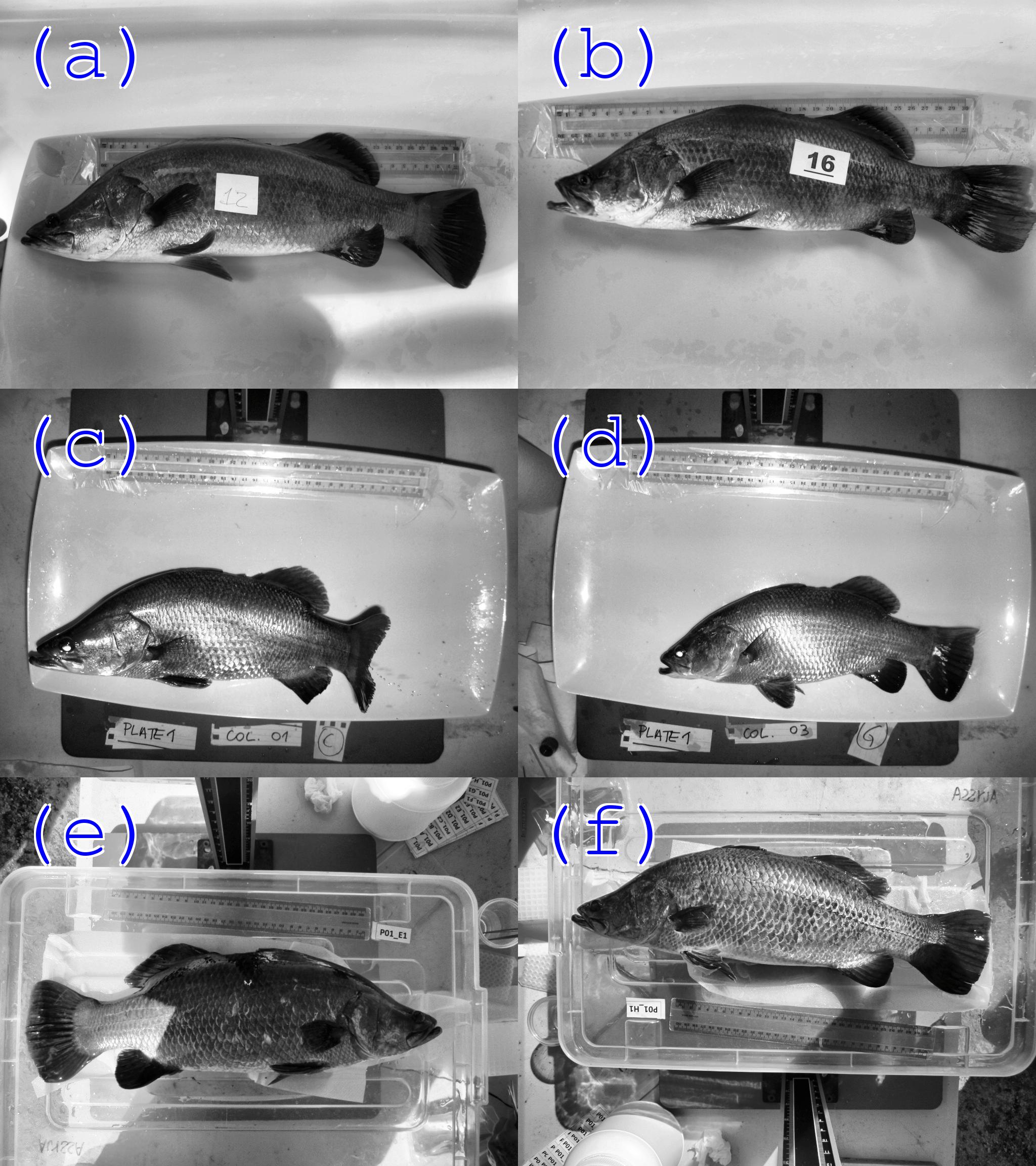}
% {./figures/fig2dk.eps}
\end{center}
\caption{The same samples as in Fig. \ref{fig:fig1} converted to grayscale 
and enhanced by CLAHE \cite{zuiderveld1994contrast}.}
\label{fig:fig2}
\end{figure}

The focus of this study was to continue developing methods for the
automatic estimation of harvested fish weight from images.
Specifically, the following practical and theoretical questions 
were addressed.
Firstly, a practical application question: is it correct to assume (e.g. in \cite{konovalov2018estimating}) 
that a model with excluded fins and tail would be more accurate 
compared to a model which used the whole fish silhouette? 
It is clearly much easier to extract the whole fish surface area than to exclude the non-exactly defined fins \cite{balaban2010prediction,viazzi2015automatic}. 
Therefore, the additional complication of using the modern Deep Convolutional Neural Networks (CNNs) \cite{konovalov2018estimating,lecun2015deep} must be justified, for example, by a significantly more accurate mass-estimation model. 
The second practical question was to test how stable the Eqs.~\ref{eq:1} and \ref{eq:2} 
models were when applied to a 
different set of barramundi images. 
From a theoretical point of view, the utilized (in \cite{konovalov2018estimating}) semantic segmentation FCN-8s CNN \cite{Shelhamer_2017,long2015fully,ronneberger2015u} 
model was replaced here by the more recent LinkNet-34 \cite{chaurasia2017linknet,shvets2018automatic} 
CNN to test the stability and accuracy of the automatically segmented with-fins and without-fins fish surface areas.

The presented weight estimation pipeline was designed to be fast enough to process video 
frames as individual images in real time for the frame 
sizes up to $480 \times 480$ resolution. 
In the aquaculture industry, a typical conveyor could be equipped with a video camera providing
a video feed for the weight estimation processing. Furthermore, 
conveyor harvesting or transporting videos could be processed off-site, making the 
estimation procedure 
more financially viable and/or more accurate by processing the frames at 
higher resolutions. The required calibration could be easily achieved by sliding or placing 
a measuring ruler (or a known size object) on the conveyor.
Note that for an actual industrial deployment
it would be required to deal with tracking of individual fish and 
having multiple fish in the same frame, which was deemed outside the scope of this study.

\section{MATERIALS AND METHODS}
\label{subsec:MATERIAL}

\subsection{Datasets}
\label{subsec:Datasets}
 Three datasets originated from \cite{domingos2014fate} were used in this study. The Barra-Ruler-445 (BR445) dataset contained 445 images with manually measured weights in the range of $1-2.5$kg. 
 BR445 was used in 
 \cite{konovalov2018estimating,konovalov2018automatic,konovalov2017ruler}, 
 see two typical examples in Figs. 1(a) and 1(b). 
 The second dataset was Barra-Area-600 (BA600)  containing more than 600 image-weight pairs (used in \cite{konovalov2018estimating}), 
 where BA600 fish weights were between 0.2~kg and 1~kg, see two examples in Figs. \ref{fig:fig1}(e) and \ref{fig:fig1}(f). 
 
The third dataset (denoted BW1400) contained 1,400 harvested barramundi images with 
corresponding weight values from the $0.15-1.0$kg range. 
The BR445 and BA600 images were taken outdoors under the natural sunlight, while BW1400 images were taken indoors under artificial lighting. Note the same white holding plate (Figs 1a-d) had a blue tint in the BR445 images (Figs. \ref{fig:fig1}a-b). To minimize dependency on 
such transient colors, in training and testing, all images were transformed to 
grayscale (Fig. \ref{fig:fig2}).

\subsection{Semantic Segmentation of Images }
\label{section:Segmentation}
\begin{figure}[!ht]
\begin{center}
\includegraphics[width=0.48\textwidth]
{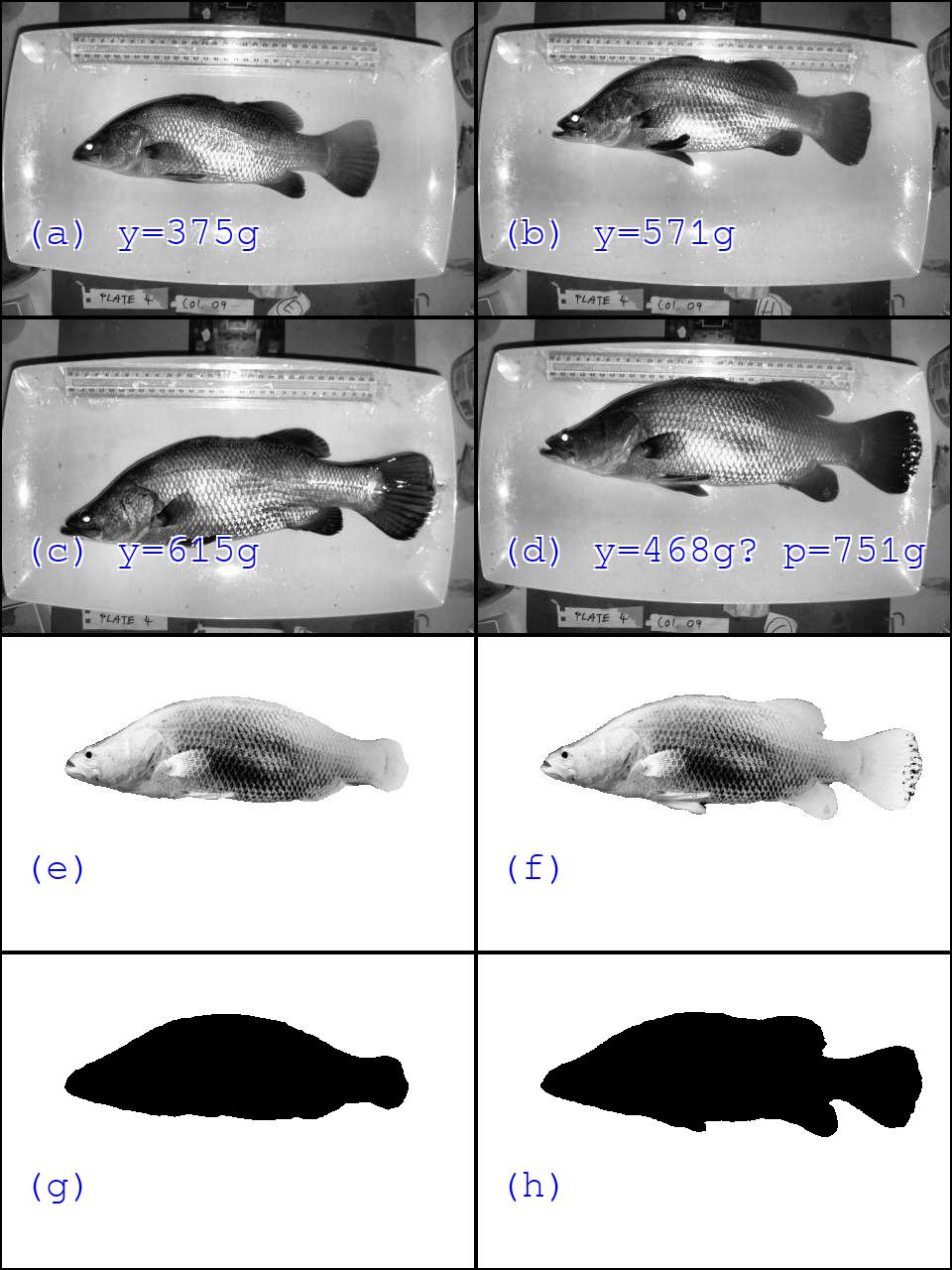}
% {./figures/fig4dk.eps}
\end{center}
\caption{An example of weight measuring error in the BW1400 dataset: 
(a-c) the correctly measured reference images with $y$ weight values; 
(d) the identified recording/measuring error (predicted $p=751$g);
(e) the mask without fins and tail for the fish in (d); 
(f) the whole fish mask for the fish in (d); 
(g) the mask without fins and tail; 
(h) the whole fish mask. 
Reversed grayscale was used in (e)-(h) for clarity.}
\label{fig:error_example}
\end{figure}

The 200 no-fins masks from \cite{konovalov2018estimating} together with the corresponding fish images were scaled to 1~mm-per-pixel, where 100 mask-image pairs were from BR445 and 100 from BA600, see examples 
in Fig.~\ref{fig:error_example}. In order to examine the fins/no-fins effect, additional 100 with-fins masks were manually segmented (50 from each BR445 and BA600), 
see example in Fig.~\ref{fig:error_example}(h). 
The lower number of the whole-fish masks (with fins) was justified by expecting the whole-fish segmentation to be a much easier problem to solve.

The most accurate Fully Convolutional Network from \cite{Shelhamer_2017}, FCN-8s, was trained on the 200 no-fins masks and applied in \cite{konovalov2018estimating}. 
Even though FCN-8s was a major theoretical breakthrough when it was reported \cite{Shelhamer_2017,long2015fully}, at the moment, FCN-8s is often less accurate than the more recent U-Net \cite{ronneberger2015u} type of segmentation CNNs. Furthermore, since only 200 no-fins masks out of the 1072 images in \cite{konovalov2018estimating} were manually segmented, it was not possible to assess the actual accuracy of FCN-8s segmentations 
on the remaining not segmented images.
Therefore, by using a different and more accurate (at least in theory) segmentation CNN in this study, we were aiming to assess the accuracy of 
the originally reported results obtained via FCN-8s. 

A variation of U-Net \cite{ronneberger2015u}, LinkNet-34 
\cite{chaurasia2017linknet}, was selected for this study, where ResNet-34 \cite{he2016deep} was used as the feature encoder and 
the PyTorch implementation was from \cite{shvets2018automatic}. Two factors contributed to the choice of LinkNet-34. First, reproducibility of CNN results remains a challenge in many cases. 
This concern was mitigated by using the standard 
ResNet-34 CNN (available in the PyTorch distribution)
together with the relatively simple LinkNet-34-style decoder, which was also available as an "off-the-shelf" downloadable component \cite{shvets2018automatic}.
The second deciding factor was that LinkNet-34 delivered a good balance of speed 
(verified during this project)
and very high accuracy, which was demonstrated in 
the MICCAI 2017 Endoscopic Vision Sub-Challenge: Robotic Instrument Segmentation
 \cite{MICCAI2017,shvets2018automatic}. 

\subsection{Training Pipeline }
\label{subsec:Pipeline}
 The training pipeline of \cite{konovalov2018estimating} was retained as much as possible, where the following steps were similar or identical to \cite{konovalov2018estimating}:
\begin{itemize}
\item The 200 no-fins masks and the 100 with-fins masks were split into 
training and validation sets as 80\% and 20\%, correspondingly.

\item ResNet-34 layers were loaded with their 
ImageNet \cite{russakovsky2015imagenet} trained weights to speed up the training process via the knowledge transfer \cite{oquab2014learning}. The \textit{sigmoid} activation function was used in the last output layer.

\item Weight decay was set to $1 \times 10^{-4}$ and applied to all trainable weights.
\item All images and masks were scaled to 1~mm-per-pixel.
\item To reduce overfitting for both training and validation, 
the image-mask pairs were randomly: 
\begin{itemize}
\item 	rotated in the range of $\pm 180$ degrees;
\item 	scaled in the range of [0.8, 1.2]; 
\item 	cropped to $480 \times 480$ pixels; 
\item   flipped horizontally and/or vertically with 0.5 probability.
\end{itemize}
\item Training was done in batches of 8 image-mask pairs.
\item Adam \cite{kingma2014adam} was used as a training optimizer.
\end{itemize}

Compared to \cite{konovalov2018estimating}, the following training steps were improved. As per \cite{shvets2018automatic}, the loss function (Eq.~\ref{eq:3}) was replaced by (Eq.~\ref{eq:4}):
\begin{equation}
\label{eq:3}
loss(y, \hat{y})=b c(y, \hat{y})+(1-dice(y, \hat{y}))
\end{equation}
\begin{equation}
\label{eq:4}
loss(y, \hat{y})=b c(y, \hat{y})-\ln (dice(y, \hat{y}))
\end{equation}
where $y$ was a target mask, $\hat{y}$ was the corresponding LinkNet34 output, $b c(y, \hat{y})$ was the binary cross entropy, $dice(y, \hat{y})$ was the Dice coefficient \cite{dice1945measures}. 
For both training and testing, the input images were converted to one-channel gray images and normalized to the [0,1] range of numerical values. 
In order to reuse the ImageNet-trained ResNet-34 encoder, additional gray-to-color trainable conversion layer was added to the front of LinkNet-34 as per \cite{konovalov2018situ}. 
In addition to the original augmentations \cite{konovalov2018estimating}, 
image blurring (kernel sizes 3 or 5 pixels) or CLAHE \cite{zuiderveld1994contrast} were applied with 0.5 probability each.

\begin{figure}[htbp]
\begin{center}
\includegraphics[width=0.48\textwidth]
% {./figures/fig_training_200.eps}
{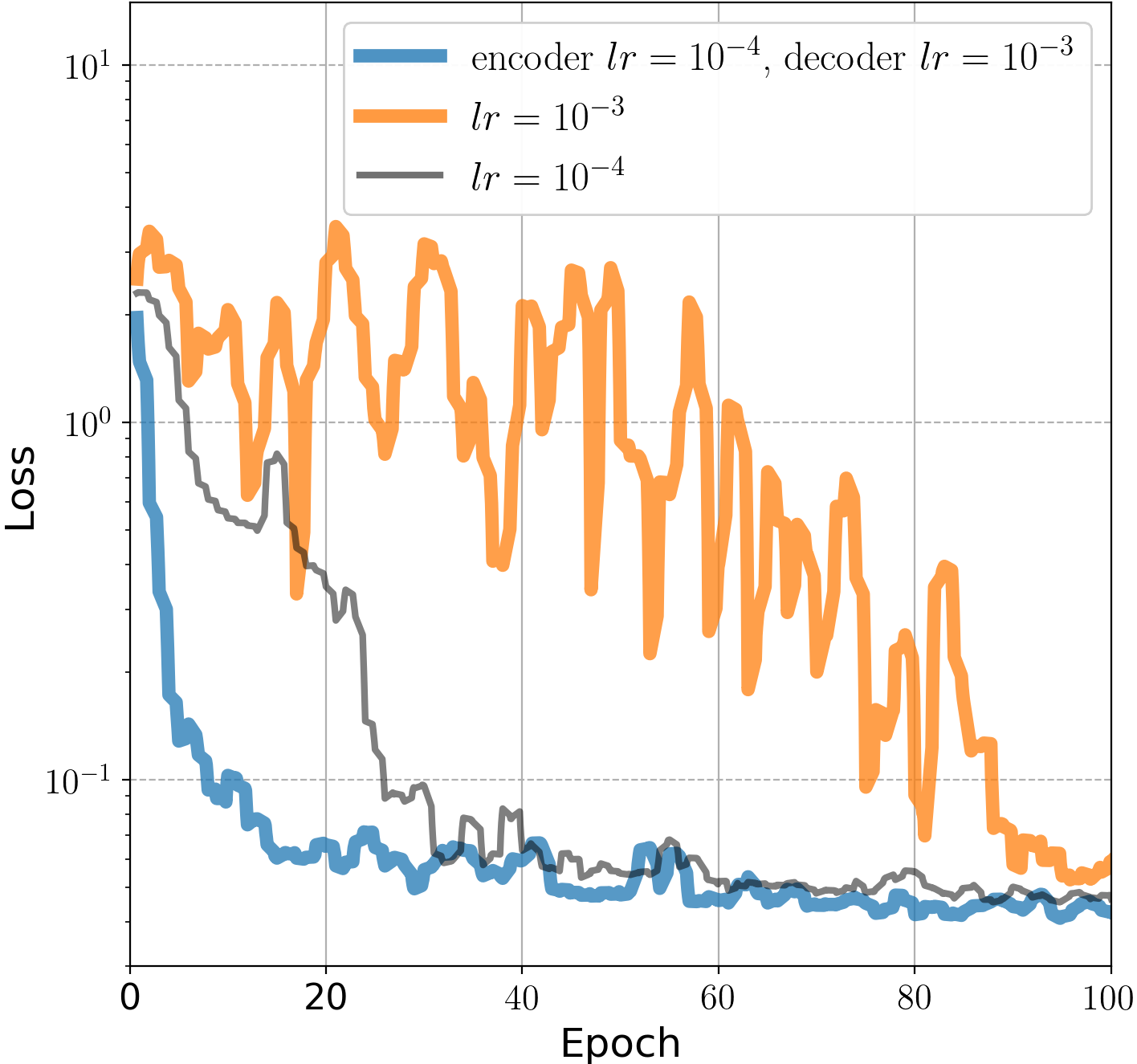}
\end{center}
\caption{Effect of learning rates on validation losses}
\label{fig:fig_train}
\end{figure}

Use of LinkNet-34 as a more advanced segmentation CNN (compared to FCN-8s) together with grayscale images (and extensive augmentations) removed the necessity of freezing the ImageNet-trained encoder weights \cite{konovalov2018estimating}. However, to assist in more effective re-use of the pre-trained ResNet-34, the Adam's learning rate was reduced by factor of 10 when applied to the encoder (ResNet-34) layers. 
Adam's starting learning rate ($l r$) was set to $l r=1 \times 10^{-3}$ and then linearly reduced by 100 to $l r=1 \times 10^{-5}$ over 100 training epochs. 
The blue line in Fig.~\ref{fig:fig_train} corresponds to the
validation loss values while training over 100 epochs.
With the same linear learning rate annealing schedule, 
if the same starting learning rate 
($l r=1 \times 10^{-3}$ or $l r=1 \times 10^{-4}$) was applied to the ImageNet trained encoder (ResNet34) and 
the randomly initialized LinkNet34 decoder layers, the validation 
loss decreased less rapidly compared to our approach, 
see Fig.~\ref{fig:fig_train}. 
If not frozen, lower learning rate was needed \cite{FastAI} for 
the ImageNet trained layers (e.g. ResNet-34 encoder) 
not to be randomized while training
together with randomly initialized layers 
(e.g. LinkNet-34 decoder layers). 

\section{Results and Discussion}
\label{sec:results}
The main goal of this study was to continue developing 
the best practice approaches to 
automatic estimation of the weight of harvested fish from images. 
This goal was approached via 
{\em weight-from-area} and {\em weight-from-image} models.

\subsection{Weight-from-area mathematical models}
The first step was to examine 
whether the simple mathematical models estimating fish mass $M$ 
from its image surface area $S$, see Eqs.~\ref{eq:1} and \ref{eq:2}, 
were accurate and reliable for the industry. 
This approach was a simple (to understand and explain)
way of any object weight estimation from its image surface area.

\subsubsection{With or without fins}
We examined two mathematical models, see Eqs.~\ref{eq:1} and \ref{eq:2}, 
if they were more accurate by using only the fish body (the “\textit{no-fins}“ rows of Table \ref{tab:TABLE1}) rather than the whole fish including fins and tail 
(the “\textit{whole}” rows in Table~\ref{tab:TABLE1}). 
The results in rows 1 and 2 (cells highlighted blue) 
revealed that for the one-factor model (Eq.~\ref{eq:1}), the coefficient of determination ($R^2$) and the mean absolute percentage error (MAPE) were indeed better for the no-fins models. Similar, the two-factor model (Eq.~\ref{eq:2})
was more accurate for the no-fins automatically segmented masks (cells highlighted orange). Here only the fitting performance of the models 
was considered, where the predictive accuracy was discussed 
in due course below.

\begin{table}[htbp]
\renewcommand{\arraystretch}{1.3}
\centering
\caption{Mass estimation models}
\label{tab:TABLE1}
\begin{tabular}{l|c|c|c|c}
\hline
\hline
{\bf Mask} & {\bf Model}  & {\bf Fit} & {\bf Fit} & {\bf BW1400} \\
{\bf type} &  Fitted or trained on & $R^{2}$ & {\em MAPE} & {\em MAPE} \\
 & BR445 and BA600  &  & [\%] & [\%] \\
\hline

{\em 1. whole} 
& $c=0.1254$ & \cellcolor[HTML]{A4BBFF}\textbf{0.976} & \cellcolor[HTML]{A4BBFF}\textbf{5.44} & 4.36 \\ 
 
{\em 2. no-fins} & $c=0.1718$
& \cellcolor[HTML]{A4BBFF}\textbf{0.979} & \cellcolor[HTML]{A4BBFF}\textbf{5.32} & 6.75 \\ 

% {\em xx. no-fins}  & $c=0.1660$, mask from \cite{konovalov2018estimating}
% & 0.980 & 5.26 & 9.69 \\ 

 & Eq.~\ref{eq:1}, log-MSE fit  &  & &  \\  

\hline
{\em 3. no-fins}  & $c=0.1702$ & 0.983 & 5.58 & 7.57 \\ 

 & Eq.~\ref{eq:1}, MSE fit \cite{konovalov2018estimating}
 &  &  & \\ 

\hline
{\em 4. whole} & $a=0.0837$, $b=1.567$
& \cellcolor[HTML]{FFCE93}\textbf{0.979} &  \cellcolor[HTML]{FFCE93}\textbf{4.68} & 6.19 \\ 

{\em 5. no-fins} 
& $a=0.1099$, $b=1.577$
& \cellcolor[HTML]{FFCE93}\textbf{0.982} & \cellcolor[HTML]{FFCE93}\textbf{4.33} & 10.35 \\ 

 & Eq.~\ref{eq:2}, log-RANSAC fit  &  &  & \\

\hline
{\em 6. no-fins} & $a=0.1239$, $b=1.550$ & 0.983 & 4.53 & 11.51 \\
% {\em 6-b. no-fins} & $a=0.1239$, $b=1.550$ & 0.983 & 4.53 & 11.51 \\
& Eq.~\ref{eq:2}, MSE fit \cite{konovalov2018estimating} 
&  &  & \\

\hline
{\em 7. whole} & LinkNet-34R &  & 4.27 & 11.4 \\ 

{\em 8. no-fins} & LinkNet-34R &  & 4.20 & {\bf 4.28} \\ 
% & Trained on 80\% of data &  &  &   \\ 
\hline
\hline
\end{tabular}
\end{table}

\subsubsection{Linear fit in logarithmic scale}
\begin{figure*}[htbp]
\begin{center}
\includegraphics[width=0.70\textwidth]
{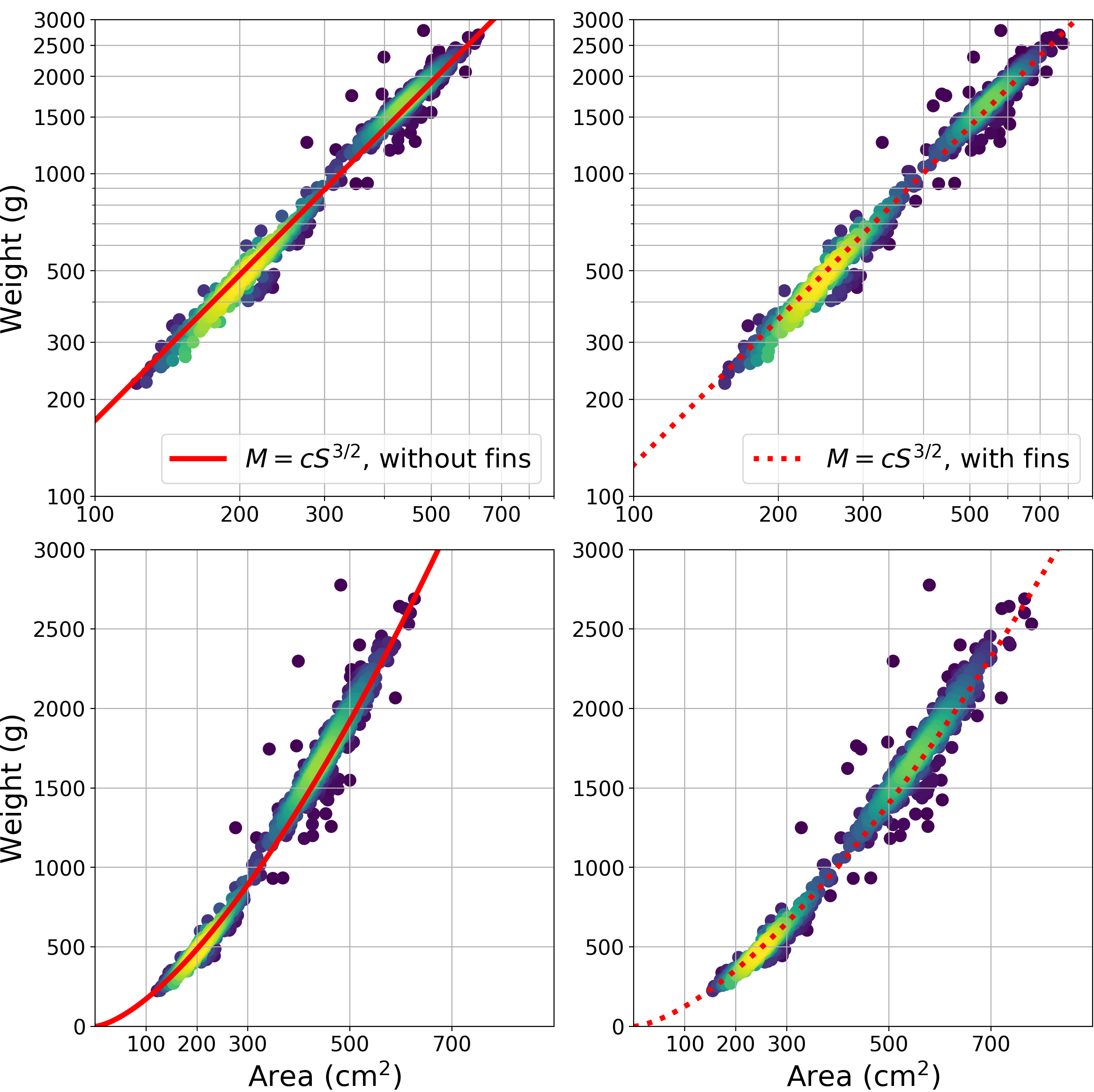}
% {./figures/fig3dk2.eps}
\end{center}
\caption{Relation between the measured fish weight and the automatically segmented fish area for the combined BR445 and BA600 datasets: 
without fins and tail (left figures), the whole fish (right figures).}
\label{fig:fig3}
\end{figure*}

Furthermore, the original fit (row 3 of Table~\ref{tab:TABLE1}) was not done in logarithmic scale and therefore larger weights  
had disproportionately larger contribution to the fit 
(compare top and bottom rows in Fig.~\ref{fig:fig3}),
which was done by minimizing the mean squared error (MSE). In this 
study, rows 1 and 2 were fitted (still by minimizing MSE) 
in logarithmic scale \cite{viazzi2015automatic}
(top row in Fig.~\ref{fig:fig3}) 
thus improving the MAPE
from 5.58\% to 5.32\% (rows 2 and 3 of Table~\ref{tab:TABLE1}).
The MAPE improvement (due to the exclusion of fins and tail) 
was less pronounced if the MSE fitting was 
done directly on the area and weight values, not their logarithms.
Fig.~\ref{fig:fig3} illustrates how qualitatively similar the without-fins (left sub-figures) and with-fins (right sub-figures) distributions were, 
where higher density of data points was drawn by a lighter (yellow) color. 
Fig.~\ref{fig:fig3} suggested a possible explanation of why
some previous studies did not detect the improvement 
from the no-fins masks \cite{balaban2010prediction}.
 
\subsubsection{Outliers and robust fit}
Fig.~\ref{fig:fig3} also exposed a number of outliers. 
One approach dealing with the possible outliers was to use robust linear regression 
\cite{KonovalovRobust2008}, which was adopted in this work by fitting 
the two-factor model (Eq.~\ref{eq:2}) via
the RANSAC algorithm \cite{RAMSAC81} in the logarithmic scale 
(top row of Fig.~\ref{fig:fig3}), see 
rows 4 and 5 in Table~\ref{tab:TABLE1}.
The two-factor fitting coefficient $b$  was
different by less than 1\%
between the with-fins ($b=1.567$) and no-fins ($b=1.577$) models
indirectly confirming that the RANSAC fit was indeed robust
to the outliers. As expected, 
the robust fit of automatically 
segmented fish silhouettes without fins and tails achieved the best 
$\mbox{MAPE}=4.33\%$ of all considered mathematical models.

\subsubsection{Image collection procedure}
Similar to the one-factor model,
an improvement of about 0.35\% was observed in the no-fins
$\mbox{MAPE}=4.33\%$, see Table \ref{tab:TABLE1} cells highlighted orange. 
However, 
the image scales were accurate to approximately 1-2\%, 
where the scales were taken from the rulers present in every image. 
The visual distortion of the ruler often yielded up to 1\% different number of pixels between the top and bottom graduation markings (per ruler length). 
Therefore, in practical sense, 
a better image collection procedure could be more important than 
excluding fins and tail for the model building purposes.

\subsection{Weight-from-image estimation}
In the preceding sections, a fish image was segmented 
into the background zero-values pixels and 
the value-of-one fish-mask with or without fins via the LinkNet-34 
segmentation CNN. The threshold of accepting the LinkNet-34 
sigmoid output as one (foreground pixels)
was not fine-tuned and was left at its default 0.5 value.
Then the total number of nonzero pixels
were added to obtain the fish area $S$, which was fitted to the 
corresponding fish weight $M$ via Eqs. \ref{eq:1} or \ref{eq:2}. 
Effectively, every foreground 
fish pixel was assumed to contribute equally to 
the total fish mass.
While the simple mathematical models were easy to interpret,
Standley {\em et al.} \cite{standley2017image2mass}, in 2017, 
reported one of the 
first applications of CNNs for image-to-mass conversion achieving 
$\mbox{MAPE} < 1\%$ on more than 1,300 test images of generic 
everyday-life and household objects, 
where the training collection had around 150,000 images.
Hence, it was interesting to explore the direct conversion of the 
segmented mask to weight via 
the regression version of  LinkNet-34, denoted LinkNet-34R.

The LinkNet-34R was obtained from LinkNet-34 by 
adding up all the LinkNet-34 sigmoid 
outputs without thresholding 
and converting the sum $y_s$ to the logarithmic scale:
\begin{equation}
\label{eq:regr_log}
    y_r = log(y_s+1),
\end{equation}
where 1 was added to assign a zero mass value
to images without detected fish foreground masks.
The automatically segmented fish images (not just masks), 
see examples in 
Fig.~\ref{fig:error_example}(e) 
and ~\ref{fig:error_example}(f), were used as inputs
to LinkNet-34R
to make sure 
that predicted weight values from the CNN outputs were
correlated to the fish image (with or without fins) versions
and not anything else.
The corresponding training fish weights were log 
scaled via the same Eq.~\ref{eq:regr_log} by  
replacing $y_s$ with the $M$ weight values. 
The LinkNet-34R training pipeline remained identical to that of 
LinkNet-34 with the only difference of not randomly 
rescaling the images,
while the random scaling within 80\%-120\% range was used for
LinkNet-34 but not for LinkNet-34R. Since the LinkNet-34 was already 
trained to detect the fish correctly, 
the LinkNet-34R version 
was loaded with the LinkNet-34 parameters and then trained
starting from the learning rates reduced by factor of 10 
in fine-tuning regime.

While running numerical experiments, 
large errors were examined and in
approximately 1-2\% of all image-weight pairs 
some image and/or recording/measuring 
errors were identified. 
For example, comparing 
identically scaled (1~mm-per-pixel) images in 
Fig.~\ref{fig:error_example}(a)-(d), 
the expected weight of the (d) case 
should be more that 615g and was predicted as 751g, while 
due to record-taking or measuring error it was recorded as 
468g.
Such obvious errors were removed from the BW1400 dataset but not 
from the BR445 and BA600 datasets, so that the results of this
study could be directly compared to 
those of \cite{konovalov2018estimating}.
As per the {\em image2mass} study \cite{standley2017image2mass} and
since quite a few outliers remained in the BR445 and BA600 datasets 
(Fig.~\ref{fig:fig3}),
Mean Absolute Error (MAE) metric 
was used as the
loss function when training the regression LinkNet-34R model.
Using MSE would have resulted in fitting the 
outliers \cite{KonovalovRobust2008}.
All 1,072 available BR445 and BA600 segmented image-weight pairs were 
randomly split into 
80\% and 20\% for training and validation subsets, respectively, 
and the training subset was used to train the LinkNet-34R models.
The validation (not training) MAPE values (4.27\% and 4.20\%) were reported in 
rows 7 and 8 of Table~\ref{tab:TABLE1}.

\begin{figure}[htbp]
\begin{center}
\includegraphics[width=0.40\textwidth]
{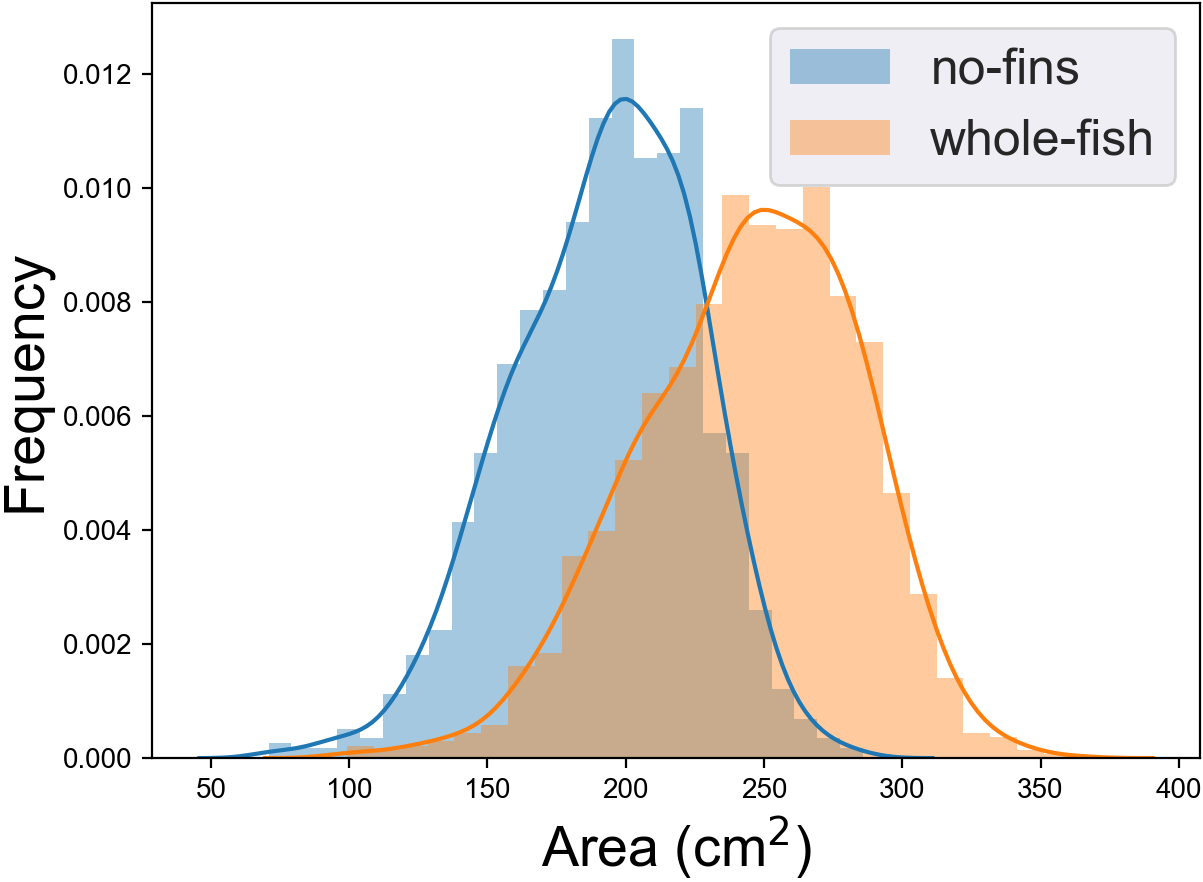}
% {./figures/fig_comp_areas.eps}
\end{center}
\caption{Normalized distributions of automatically segmented 
mask areas in the BW1400 images.}
\label{fig:comp-areas}
\end{figure}

\subsection{Predictive performance of the models}

As such, fitting known fish weights via a mathematical model or
a neural network has little practical value unless the models
could predict fish weights from new fish images. 
The last column of Table~\ref{tab:TABLE1} examined the 
predictive accuracy of the models, which were fitted 
on BR445 and BA600 and then applied to the new BW1400 dataset.
In practical industrial applications, 
the theoretical metrics such as $R^2$ becomes largely irrelevant,
hence only MAPE was discussed hereafter.
Our interpretation of the somewhat contradictory
MAPE values were as follows. 

\subsubsection{Whole-fish mathematical models predicted better}
In row 1 of Table~\ref{tab:TABLE1}, 
it was unrealistic to accept that the test $\mbox{MAPE}=4.36\%$ was indeed
better than the fitting $\mbox{MAPE}=5.44\%$.
Nevertheless, both the one- and two-factor {\em whole-fish} models 
achieved significantly better MAPE values (4.36\% and 6.19\%)
for the unseen BW1400 images, compared to the corresponding no-fins models (6.75\% and 10.35\%). 
This was consistent with the no-fins models 
from \cite{konovalov2018estimating}, rows 3 and 6 in Table~\ref{tab:TABLE1}, 
when applied to the new data.

\subsubsection{Errors in no-fins masks had larger effect}
Trying to understand why the whole-fish models predicted better, it was noticed that very often the lower-front (pelvic) fins overlapped the body and were segmented out by the no-fins CNN, see examples in Figs.~\ref{fig:fig2}(c), \ref{fig:fig2}(d) 
and \ref{fig:error_example}(b). On average, 
the no-fins mask areas were 20\% smaller than the corresponding
whole-fish areas, see Fig.~\ref{fig:comp-areas}. Therefore, the 
erroneous reductions of no-fins masks 
(e.g. due to the overlapping pelvic fins)
had larger weight error contributions 
than the variations of the fins in the
whole-fish masks.

\subsubsection{Two-factor models overfitted and
one-factor models predicted better}
Further insight was gained by observing how the 
one-factor models
(4.36\% and 6.75\% MAPEs, rows 1 and 2 in Table~\ref{tab:TABLE1})
performed much better than the corresponding
two-factor models (6.19\% and 10.35\% MAPEs, row 4 and 5). 
Therefore, the better fitting performance of 
the two-factor models (4.48\% and 4.33\%, rows 4 and 5) was 
most likely just the overfitting of the training datasets,
which was consistent with the one-factor model 
remained more stable when refitted on all available 
BR445 and BA600 samples
in \cite{konovalov2018estimating}.

\subsubsection{Direct weight-from-image CNN regression}
The simple mathematical models (Eqs.~\ref{eq:1} and~\ref{eq:2}) were based on the 
hypothesis that each fish pixel contributed equally to the total fish
weight. The preceding results indicated that
the hypothesis could be a very crude approximation, which did not perform
well beyond the one-factor models.
By forgoing the easy interpretability of the Eqs.~\ref{eq:1} and~\ref{eq:2},
the LinkNet-34R CNN models performed highly 
non-linear conversion of the segmented fish images to weights. 
The no-fins version achieved nearly identical 
validation $\mbox{MAPE}=4.20\%$ and test $\mbox{MAPE}=4.28\%$, 
see row 8 in Table~\ref{tab:TABLE1}.
However, the whole-fish version exhibited some 
overfitting similar
to the two-factor model: validation 
$\mbox{MAPE}=4.27\%$ but test $\mbox{MAPE}=11.4\%$
(row 7 in Table~\ref{tab:TABLE1}).

Detailed investigation of how the CNNs arrived at the weight predictions
was left for future work. 
In this study, we could only suggest the following 
speculative explanation. The no-fins fish images,
see example in Fig.~\ref{fig:error_example}(e), 
had smooth contour therefore LinkNet-34R had to use other 
features from within the fish images to calculate the weight.
The whole-fish contours, see Fig.~\ref{fig:error_example}(f),
were more complex and therefore were more likely
to be memorized for the individual training images, and
hence overfitted by the LinkNet-34R's more than 21 
million parameters.

\section{Conclusion}
\label{sec:Conclusion}
 
Estimation of object mass from images is an emerging area of 
computer vision \cite{standley2017image2mass} with potentially high 
impact industrial applications.
We demonstrated how a standard 
``{\em off-the-shelf}'' segmentation CNN like
LinkNet-34 from \cite{shvets2018automatic} 
could be trained efficiently using: (i) 
only 100-200 training image-mask pairs; (ii) a linear learning rate
annealing schedule; and (iii) 
reduced learning rate for the ImageNet-trained encoder (ResNet-34). 
With- or without-fins fish masks were automatically segmented and fitted
by simple mathematical models achieving 4-10\% 
MAPE values (mean absolute percentage errors consistent with other studies, e.g. 
\cite{sanchez2018automatic,viazzi2015automatic}) on 1,400 test 
images not used in the fitting procedure 
and from different geographical location.

The first question of this study was to assess if a fish silhouette 
automatically segmented by the CNNs should or should not include fish fins and tail.
Remarkably, the two simple mathematical 
models based on the whole-fish silhouette generalized better (lower MAPEs) 
when applied to the
unseen test images from the different geographical location.
The second main question was answered by demonstrating that the simplest one-factor 
(one-parameter) mathematical model performed better than the two-factor model 
on the new test images. Furthermore, the one-factor model was highly stable
achieving a lower $\mbox{MAPE}=4.36\%$ on the test images than on the training 
images, $\mbox{MAPE}=5.44\%$.

We successfully
tested a conversion of a segmentation CNN, LinkNet-34,
to weight-predicting CNN, LinkNet-34R, achieving 4-11\% test MAPE valuess.
To the best of our knowledge, this study presents the first 
practical and easily reproducible 
weight-from-image approach, e.g. by downloading the LinkNet-34 CNN
together with the corresponding 
training pipeline from \cite{shvets2018automatic} and then following
the steps explained in this study.
However, only the no-fins version of the direct regression via LinkNet-34R 
performed well on the test images strongly indicating possible overfitting of 
the whole-fish version. 

\section*{ACKNOWLEDGMENT}
We gratefully acknowledge the Australian Research Council Linkage Program schemes,
who funded the work that generated the datasets, and Mainstream Aquaculture, 
the industry partner.

\bibliographystyle{IEEEtran}
% argument is your BibTeX string definitions and bibliography database(s)
%\bibliography{IEEEabrv,../bib/paper}
\bibliography{IEEEabrv,FishMass2019}

%
% <OR> manually copy in the resultant .bbl file
% set second argument of \begin to the number of references
% (used to reserve space for the reference number labels box)
% \begin{thebibliography}{1}
% \bibitem{IEEEhowto:kopka}
% H.~Kopka and P.~W. Daly, \emph{A Guide to \LaTeX}, 3rd~ed.\hskip 1em plus
%   0.5em minus 0.4em\relax Harlow, England: Addison-Wesley, 1999.
% \end{thebibliography}
 
% that's all folks
\end{document}